\documentclass[10pt,twocolumn,letterpaper]{article}

\usepackage{cvpr}
\usepackage{times}
\usepackage{epsfig}
\usepackage{graphicx}
\usepackage{amsmath}
\usepackage{amssymb}
\usepackage{xcolor}

\usepackage{multirow}
\usepackage{soul}
\usepackage{pifont}
\newcommand{\cmark}{\ding{51}}%
\newcommand{\xmark}{\ding{55}}%
\usepackage{subcaption}

\DeclareMathOperator*{\argmax}{arg\,max}

\usepackage[breaklinks=true,bookmarks=false]{hyperref}

\cvprfinalcopy 

\def\cvprPaperID{4741} 

\ifcvprfinal\fi
\begin{document}

\title{\textit{How to Make a BLT Sandwich?} \\Learning to Reason towards Understanding Web Instructional Videos} 

\author{Shaojie Wang, Wentian Zhao, Ziyi Kou, Chenliang Xu\\
University of Rochester\\
{\tt\small \{shaojie.wang, wentian.zhao, ziyi.kou, chenliang.xu\}@rochester.edu}
}

\maketitle

\begin{abstract}
Understanding web instructional videos is an essential branch of video understanding in two aspects. First, most existing video methods focus on short-term actions for a-few-second-long video clips; these methods are not directly applicable to long videos. Second, unlike unconstrained long videos, e.g., movies, instructional videos are more structured in that they have step-by-step procedure constraining the understanding task. In this paper, we study reasoning on instructional videos via question-answering (QA). 
Surprisingly, it has not been an emphasis in the video community despite its rich applications. We thereby introduce YouQuek, an annotated QA dataset for instructional videos based on the recent YouCook2~\cite{Youcook}. The questions in YouQuek are not limited to cues on one frame but related to logical reasoning in the temporal dimension. Observing the lack of effective representations for modeling long videos, we propose a set of carefully designed models including a novel Recurrent Graph Convolutional Network (RGCN) that captures both temporal order and relation information. Furthermore, we study multiple modalities including description and transcripts for the purpose of boosting video understanding. Extensive experiments on YouQuek suggest that RGCN performs the best in terms of QA accuracy and a better performance is gained by introducing human annotated description. 
\end{abstract}

\section{Introduction}
\label{sec:intro}
\begin{figure}[]
    \centering
    \includegraphics[width=0.48\textwidth]{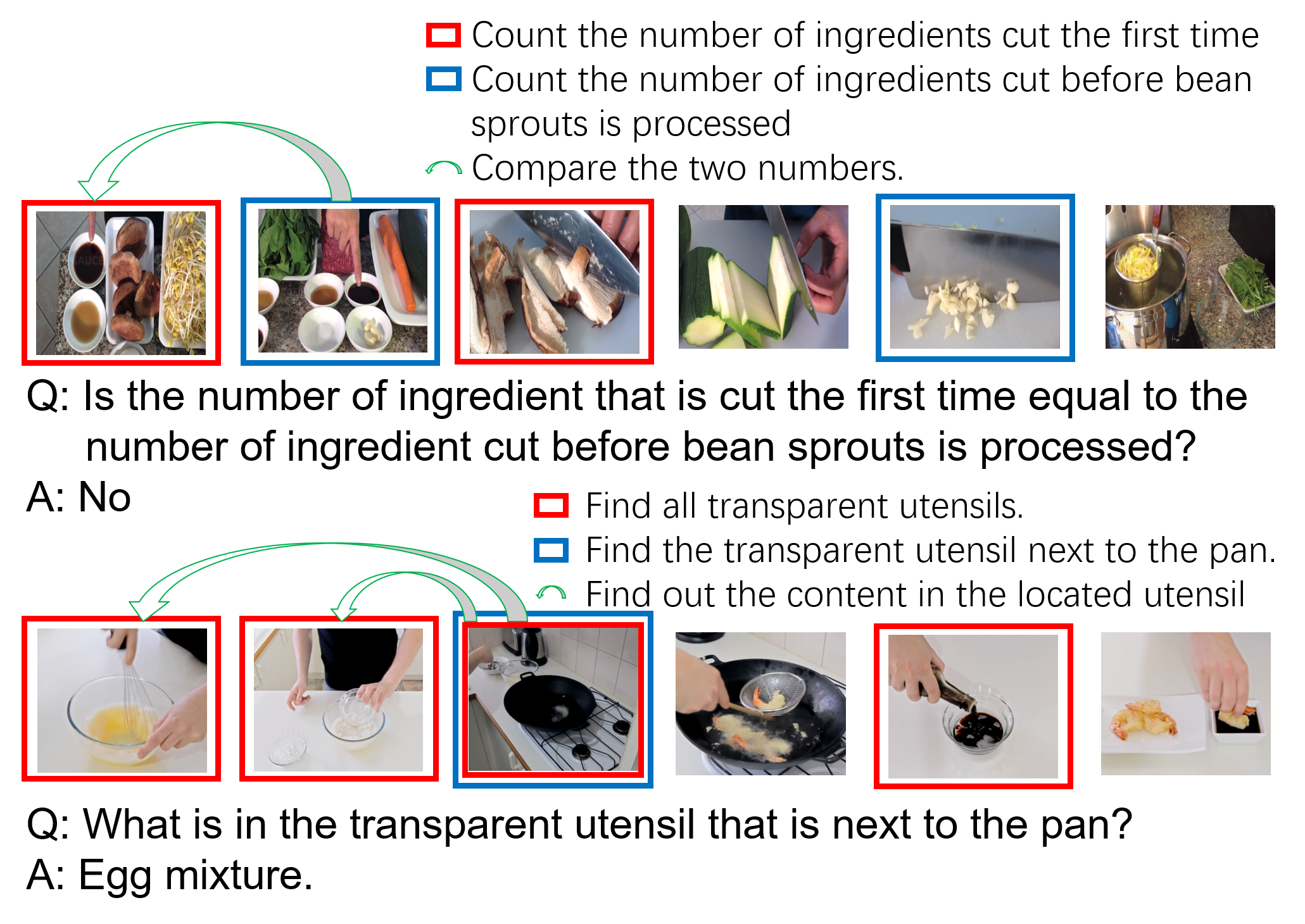}
    \caption{Demonstration of YouQuek dataset. Colored boxes and arrows represent different reasoning steps required to answer the given questions. Red boxes denote the first step, blue boxes denote the second, and green arrows are for the final step. Better view zoomed in and with color. } 
    \label{youquek}
\end{figure}

Humans can acquire knowledge by watching instructional videos online. A typical situation is that people confused by specific problems try to look for solutions in related instructional videos. For example, while learning to cook new dishes, they may wonder \textit{why a specific ingredient is added}, and \textit{what happens between the two procedures}. Watching instructional videos can often clarify these questions and hence, assist humans in accomplishing tasks. We hereby propose the question: can machines also understand instructional videos as humans do, which requires not only accurate recognition of objects, actions, and events but also the \textit{higher-order inference} of any relations therein, e.g., spatial, temporal, correlative and casual? Here we use higher-order inference to refer to the inference that cannot be completed immediately by direct observations and thus requires stronger semantics for video modeling (see Fig.~\ref{youquek}). 


Current instructional video understanding studies focus on various tasks e.g., reference resolution~\cite{visual-linguistic}, procedure localization~\cite{Youcook,changing-tires-cvpr}, dense captioning~\cite{dense-captioning,video-captioning}, activity detection~\cite{MPI,making-coffee-cvpr} and visual grounding~\cite{finding-it,visual-grounding}. Despite the rich literature and applications, question-answering (QA) task in instructional videos explored in our work is less developed, which acts as a proxy to benchmark the higher-order inference in machine intelligence. Previous works, e.g., ImageQA~\cite{image_QA_1, image_QA_2, image_QA_3} and VideoQA~\cite{MovieQA, leveraging}, also leverage the QA task as automatic evaluation method, but QA on instructional videos has never been tackled before.



Observing the lack of suitable dataset on instructional videos, we propose YouCook Question Answering (YouQuek) dataset based on YouCook2~\cite{Youcook} which is the largest instructional video dataset. Our YouQuek dataset is the first reasoning-oriented dataset aimed for instructional videos. We employ question-answering as intuitive interpretations for various styles of reasoning. Figure~\ref{youquek} presents two exemplar QA pairs in our dataset along with the corresponding example human reasoning procedure involved to answer the questions. YouQuek dataset contains 15,355 manually-collected QA pairs that are divided into different categories regarding different reasoning styles, e.g., counting, ordering, comparison, and changing of properties.

Upon the newly built dataset, we explore in two directions. The first one concerns effective representations of modeling instructional videos. The videos in our consideration have an average length of 5.27 min and as instructional videos, they are structured and have step-by-step procedure constraining the understanding task. By modeling the temporal relations among different procedures, we are expecting valuable information to be extracted from the instructional videos, for which we study various model structures and propose a novel Recurrent Graph Convolutional Network (RGCN). The RGCN deals with complex reasoning by message passing in the graph, but also maintains the sequential ordering information by a supporting RNN. In this design, graph and RNN can boost each other since the information can be swapped between the two pathways.

Second, we explore the use of different modalities in video modeling. Apart from visual information, temporal boundaries, descriptions for each procedure, and transcripts are explored. In this direction, we want to test the effect of combining various types of available annotations with our developed video models on understanding instructional videos. Given that modeling instructional videos from vision alone is hard, combining such information approximates better the human learning experiences and it, in turn, gives us a hint for devising better models for machine intelligence. 

We conduct extensive experiments on the YouQuek dataset. In the ablation study, we find that attention mechanism helps boost the performance. Our proposed RGCN model outperforms all other models with respect to the overall accuracy, even without attention. From the multi-modality perspective, modeling instructional videos using temporal boundaries together with descriptions can help dig more valuable information from videos. We also conduct human quiz on the QAs in our dataset. Results show that machines still have a large gap to human performance in that even without visual information, humans still can answer some questions correctly using life experience, or common sense, which hints us that incorporating the external knowledge with video models will be helpful for future works.

Our main contributions are summarized as follows.
\begin{itemize}
 \item We propose YouQuek dataset, the first reasoning-oriented dataset for understanding instructional videos. 
 \item We propose both models with various structures, especially a novel RGCN model, for video modeling. Our RGCN outperforms all other models even without attention.
 \item We incorporate multi-modal information to perform extensive experiments on YouQuek showing that description can boost the video understanding capability, while transcripts could not. 
\end{itemize}

The rest of the paper is organized as the following. We first discuss some related works in Sec.~\ref{sec:related}, and introduce the proposed YouQuek dataset in Sec.~\ref{sec:dataset}. Then in Sec.~\ref{sec:model}, we set up series of baseline models for the dataset, and propose RGCN as a new model for instructional video reasoning. In Sec.~\ref{sec:experiment}, we demonstrate and discuss the experiment results. Conclusions are drawn in Sec.~\ref{sec:conclusion}. \textit{The YouQuek dataset and our code for all methods will be released upon acceptance.}

\section{Related Work}
\label{sec:related}

\textbf{Instructional Video Understanding:} Instructional video understanding has received much attention recently. Alayrac et al.~\cite{changing-tires-cvpr} and Kuehne et al.~\cite{making-coffee-cvpr} both leverage the natural language annotation of the videos to learn the instructional procedure in videos. Zhou et al.~\cite{Youcook}, however, propose to learn the temporal boundaries of different steps in a supervised manner without the aid of textual information. Dense captioning is also posed on instructional videos in~\cite{dense-captioning}, which aims at localizing temporal events from a video, and describing them with natural language sentences. Visual-linguistic ambiguities can be a common problem in instructional videos with narratives. Huang et al.~\cite{visual-linguistic} focus on such ambiguities caused by the changing in visual appearance and referring expression, and aim to resolve references with no supervision. Huang et al.~\cite{finding-it} perform visual grounding task in instructional videos, also coping with visual-linguistic ambiguities. Yet, none of these works have tackled the QA problem on instructional videos, despite the uniqueness for instructional videos to perform reasoning.


\textbf{Video Question Answering:} People are gaining interests in video question answering (VideoQA) in recent years. Most of the current VideoQA tasks are focusing on direct facts in short videos~\cite{Unifying, spatio-temopral, leveraging, gradually, uncovering}. They all automatically generate QA pairs using a state-of-the-art question generation algorithm proposed in~\cite{good-question}. However, such auto-generation mechanism often generates QA pairs with poor quality and low diversity, though grammatically correct. Worse still, auto-generated QA pairs cannot involve reasoning. From the reasoning point of view, MovieQA~\cite{MovieQA} use human annotated QA pairs on movies to evaluate automatic story comprehension. SVQA~\cite{video-clevr}, following the step of~\cite{CLEVR}, extend the CLEVR dataset to the video version. Yet, it still focuses on short-term relations, and does not fit natural settings.


\section{YouQuek Dataset}
\label{sec:dataset}

To validate the proposed task on instructional video reasoning, we introduce YouQuek dataset, a reasoning-oriented video question answering dataset based on YouCook2 dataset. The dataset contains 15,355 question-answer (QA) pairs in total. Tailored for our dataset, we annotate the QA pairs with six different tags, where each QA pair could be labeled with more than one tag. In supplementary material, we show example QA pairs for each tag described below.

\noindent \textbf{Counting:} \quad
This tag annotates a QA pair that involves counting. One may count the occurrence time of certain actions or the number of certain ingredients. E.g., ``How many white ingredients are used in the recipe?'' Apart from counting, we also need to find out the target ingredients according to their colors.

\noindent \textbf{Time:} \quad 
Time is a distinguishing feature in videos compared to images. This category of questions are mainly about timing and duration. A typical example is, ``Which one is faster: adding water or adding salt?''. To answer this question, we not only need to know how long it takes for both actions, but also need to make comparison of the duration.

\noindent \textbf{Order:} \quad
Long-term temporal order is a unique feature for instructional videos, because instructional videos come with step-by-step procedures, and the order information matters. E.g., in YouCook2, the ordering of procedure is critical to the success of one recipe. Therefore, we stress out questions related to \textit{action orders}, e.g., ``What happens before/after/between ...?'', and ``Does it matter to change the order of ... and ...?''

\noindent \textbf{Taste:} \quad 
YouCook2 is an instructional cooking video dataset, so we bring up with the taste questions. This type of QA pairs is about the flavor and the texture of the dish. Taste can also be related to reasoning in that one can infer the taste from the ingredients used, and the texture from the cooking methods applied. Note that we avoid questions that are subjective such as ``Is this burger tasty?'', which cannot be answered by reasoning, but by subjective inspection.

\noindent \textbf{Complex:} \quad 
This tag presents a broader concept than all other tags above. By ``complex'', we emphasize a multi-step reasoning process instead of one-step reasoning. This type of questions overlaps with all other types.

\noindent \textbf{Property:} \quad 
Cooking usually involves changes of ingredients. The properties of ingredients, e.g. their shape, color, size, location, etc., may vary at different time points as the cooking procedure goes on. This type of questions is different from ``order'' questions since we are asking about certain ingredients rather than actions.

In Tab.~\ref{comparison}, we contrast our dataset to some other VideoQA datasets. Our dataset is unique in that we not only build the dataset based on instructional videos, but also focus on long-term ordering and higher-order inference. 

\begin{table*}
\caption{Comparison among different video question answering datasets. The first four columns are: ``Inst.'' for whether it is based on instructional videos; ``Natural'' for whether videos are of natural world settings; ``Reason'' for whether questions are related to reasoning; ``Human'' for whether QA pairs are collected through human labor.}
\begin{tabular}{l|cccc|l|l|l}
\hline
                                  & Inst.         & Natural      & Reason       & Human        & \# of QA & Per video length    & Answering form               \\ \hline
VTW~\cite{leveraging}             & \xmark       & \cmark       & \xmark       & \xmark       & 174955   & 1.5 min             & Open-ended                   \\ \hline
Xu et al.~\cite{gradually}        & \xmark       & \cmark       & \xmark       & \xmark       & 294185   & 14.07 sec           & K-Space                      \\ \hline
Zhu et al.~\cite{uncovering}      & \xmark       & \cmark       & \xmark       & \xmark       & 390744   & \textgreater 33 sec & Fill in blank                \\ \hline
Zhao et al.~\cite{spatio-temopral}& \xmark       & \cmark       & \xmark       & \xmark       & 54146    & 3.10 sec            & Open-ended                   \\ \hline
SVQA~\cite{video-clevr}           & \xmark       & \xmark       & \cmark       & \xmark       & 118680   & -                   & K-Space                      \\ \hline
MovieQA~\cite{MovieQA}            & \xmark       & \cmark       & \cmark       & \cmark       & 6462     & 200 sec             & Multiple choice              \\ \hline
YouQuek (Ours)                    & \cmark       & \cmark       & \cmark       & \cmark       & 15355    & 5.27 min            & Multiple choices and K-Space \\ \hline
\end{tabular}
\label{comparison}
\end{table*}

\subsection{QA collection}
Many existing VideoQA datasets~\cite{leveraging, gradually, spatio-temopral, Unifying, uncovering} adopt an automatic question-answer (QA) generation technique proposed by~\cite{good-question} to generate QA pairs from texts. However, QA pairs obtained via this method suffer from extremely low diversity. Also, automatic methods cannot generate questions involving complex reasoning, which goes against our goal of constructing the dataset. Therefore, we apply Amazon Mechanical Turk (AMT) to collect question and answer pairs. For details about the collection of QA and multiple choice alternatives, please refer to supplementary material.




\subsection{Statistics}
\begin{figure}
\centering
\begin{subfigure}[t]{0.23\textwidth}
\centering
\includegraphics[width=\textwidth]{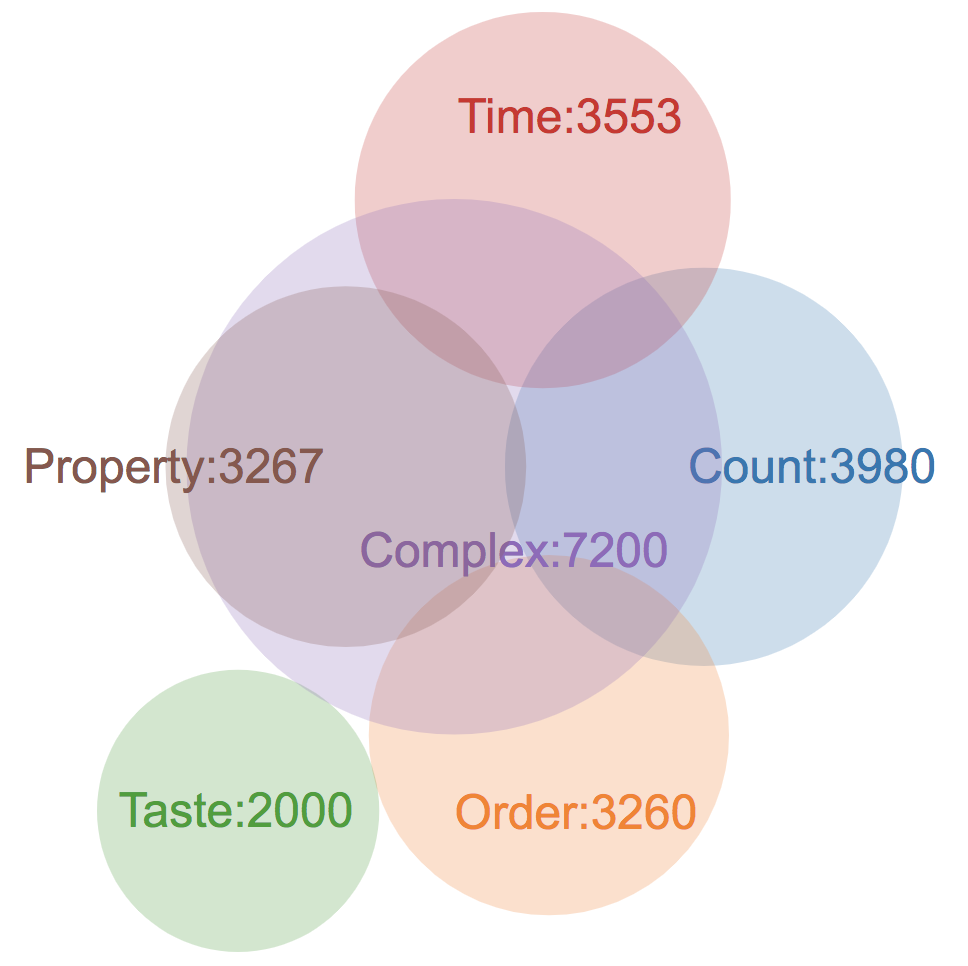}
\caption{Distribution on categories.}\label{fig:categories}
\end{subfigure}
\hspace*{\fill} 
\begin{subfigure}[t]{0.23\textwidth}
\centering
\includegraphics[width=\textwidth]{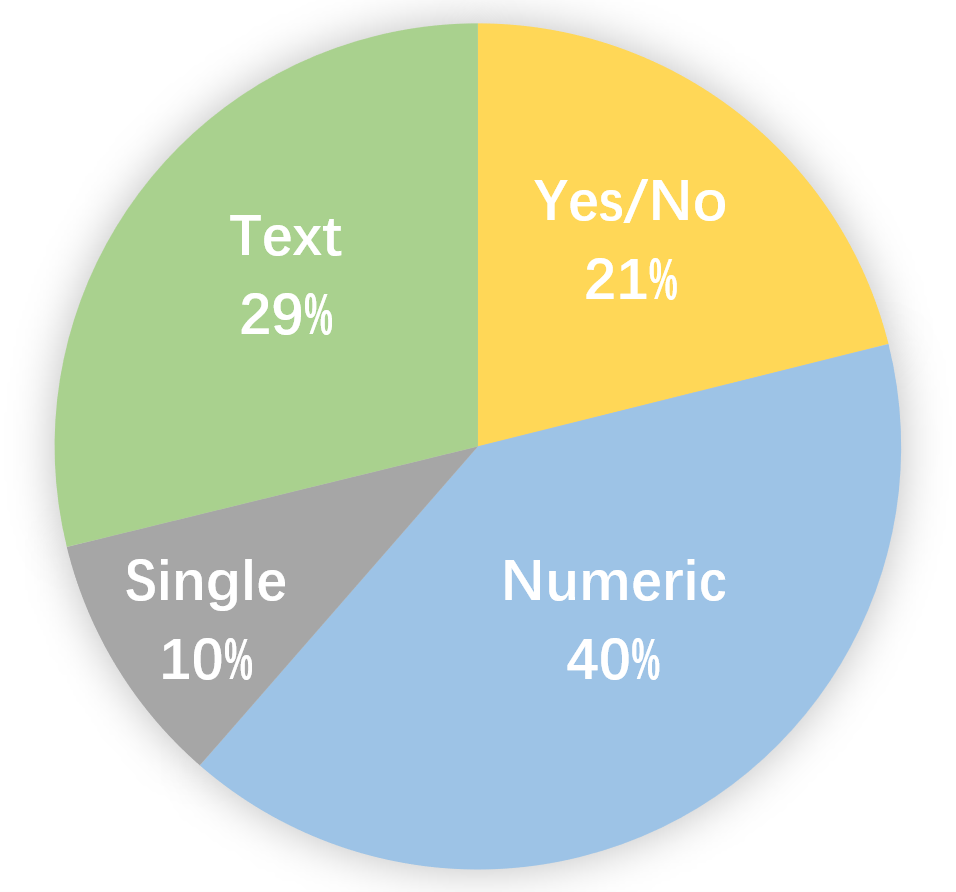}
\caption{Different answer tags.}\label{fig:tags}
\end{subfigure}
\caption{Statistics for our dataset.} \label{fig:stats} 
\end{figure}

In Fig.~\ref{fig:categories}, we show the statistics of six different categories of questions. We have 7,200 complex reasoning QA pairs, consisting nearly half of our dataset. Other questions involve simpler reasoning procedure, but still cannot be answered by direct observation from the videos. On average, we have 1.478 tags per QA pair, 2.289 words per answer, and 7.678 QA pairs per video.

To illustrate our dataset better, we split the QA pairs into four categories with respect to answer types, namely ``Yes/No'' for answers containing yes or no; ``Numeric'' for answers containing numbers, mostly related to counting and time; ``Single word'' for answers with only one word, excluding QA pairs in ``Yes/No'' and ``Numeric''; ``Text'' for answers with multiple words, excluding QA pairs in ``Yes/No'' and ``Numeric''. Fig.~\ref{fig:tags} shows the distribution of four different types of answers in our dataset.


\section{Instructional Video Reasoning}
\label{sec:model}

With the newly collected YouQuek dataset, we perform reasoning tasks by answering questions on instructional videos. We first formally define our problem in Sec.~\ref{sec:problem}. Then in Sec.~\ref{sec:models}, based on attention mechanism, we design sequential model (SEQ-SA) and graph convolutional model (GCN-SA). We also propose Recurrent Graph Convolutional Network (RGCN) which captures both temporal order and complex relations to overcome the limitation of SEQ-SA and GCN-SA. In Sec.~\ref{sec:modalities}, additional modalities such as description and transcripts are added to the reasoning model to help gain better performance.

\subsection{Problem Formalization}
\label{sec:problem}

\textbf{Multiple Choice:} Since the questions in the YouQuek dataset have alternative choices, we can use a three-way score function $f(v,q,a)$ to evaluate each alternative and choose the one with the highest score as correct answer: 
\begin{align}
j^{*}=\argmax_{j=1,\dots,M} f(v,q,a_j)
\enspace,
\label{f_vqa_mc}
\end{align}
where $M=5$ in our case, and $v$, $q$, $a$ represent the feature of video, question and answer respectively. In this work, $q$ and $a$ are the final hidden states by encoding the question and answer via RNNs. Here, $f(\cdot,\cdot,\cdot)$ denotes a MLP whose input is the concatenation of $v$, $q$, and $a$ and output is a single neuron classifying how likely the given answer $a$ is the correct one.

\textbf{K-Space:} Similar to other visual QA problems, the reasoning task can also be formulated as a classification problem on the answer space. Then the alternative (negative) answers are all other answers in the training set. Here, $K$ types of distinct answers are assigned to $K$ categories $\{A_i\}_{i=1}^K$. A MLP with $K$ output neurons is tasked to predict the correct answer $A^*$ by taking in $v$ and $q$:  
\begin{align}
A^{*}=\argmax_{j=1,\dots,K} g_j(v,q)
\enspace,
\label{f_vqa_ks}
\end{align}
where $g_j$ denotes the output score of the $j$-th neuron.

\subsection{Models}
\label{sec:models}

In this section, we mainly focus on the design of video models that can capture procedure relations in instructional events. Their generated video feature $v$ will be used for question answering. First, we describe how we pre-process the videos. Then, we introduce the architecture of proposed models that are suitable for VideoQA. Especially, we propose a novel RGCN architecture that can perform message passing between two paths: RNN and GCN, in order to capture both time series and global properties for modeling instructional videos. 

\begin{figure*}
\centering
\begin{subfigure}[t]{0.7\textwidth}
\centering
\includegraphics[width=\textwidth]{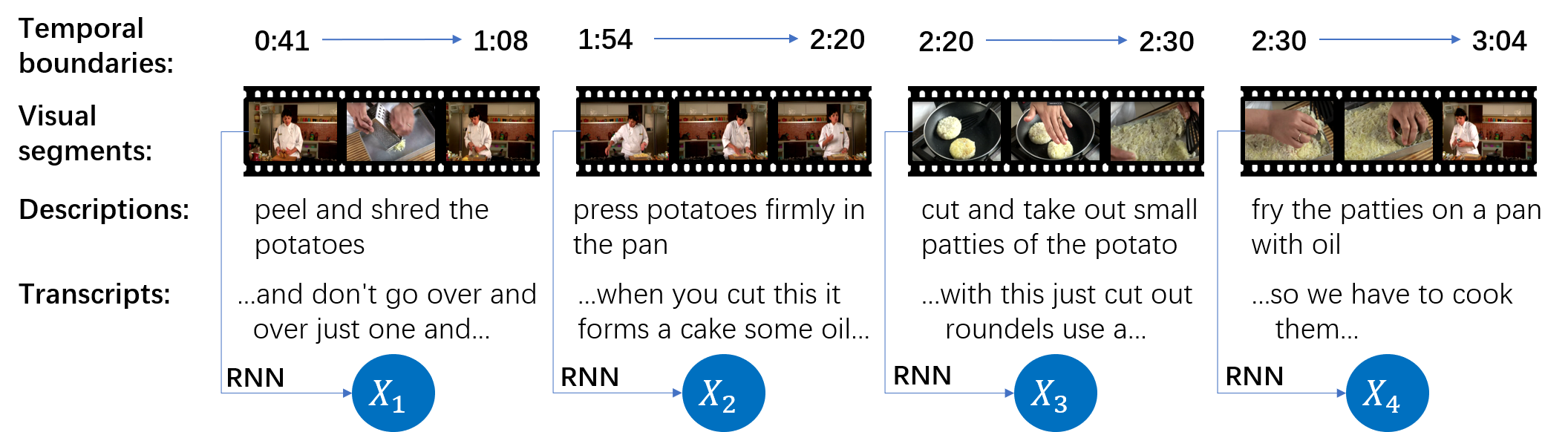}
\caption{Preprocessing}\label{preprocessing}
\end{subfigure}
\hspace*{\fill} 
\begin{subfigure}[t]{0.29\textwidth}
\centering
\includegraphics[width=\textwidth]{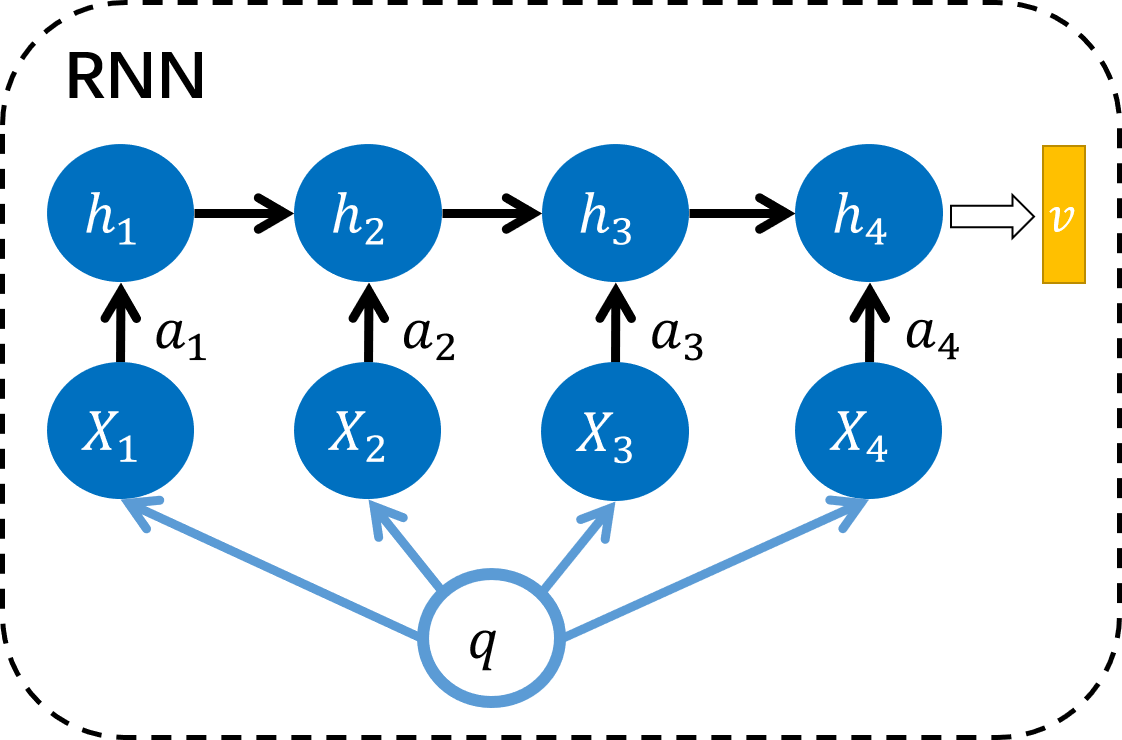}
\caption{SEQ-SA}\label{SEQ-SA}
\end{subfigure}
\hspace*{\fill} 
\begin{subfigure}[t]{0.42\textwidth}
\centering
\includegraphics[width=\textwidth]{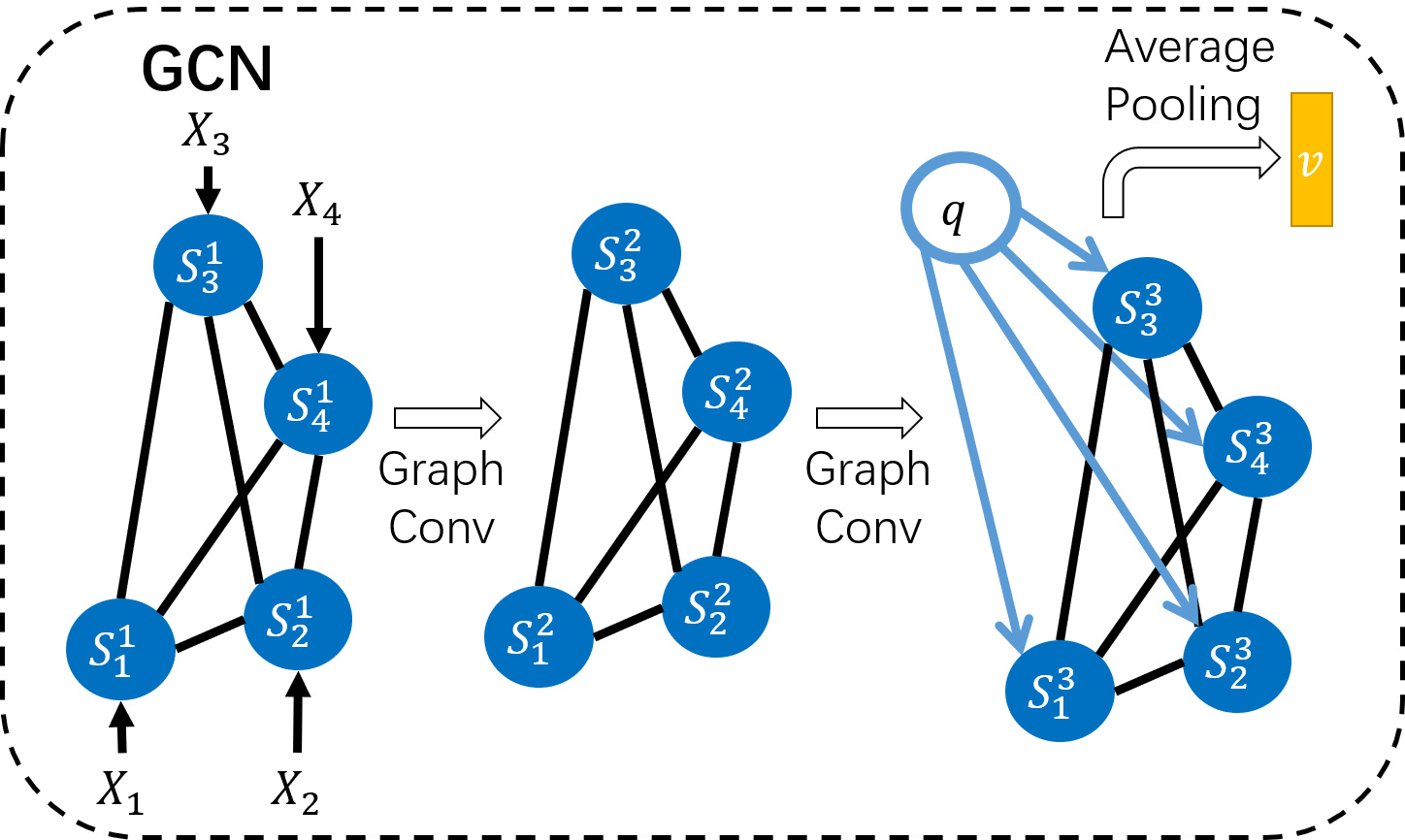}
\caption{GCN-SA}\label{GCN-SA}
\end{subfigure}
\hspace*{\fill} 
\begin{subfigure}[t]{0.56\textwidth}
\centering
\includegraphics[width=\textwidth]{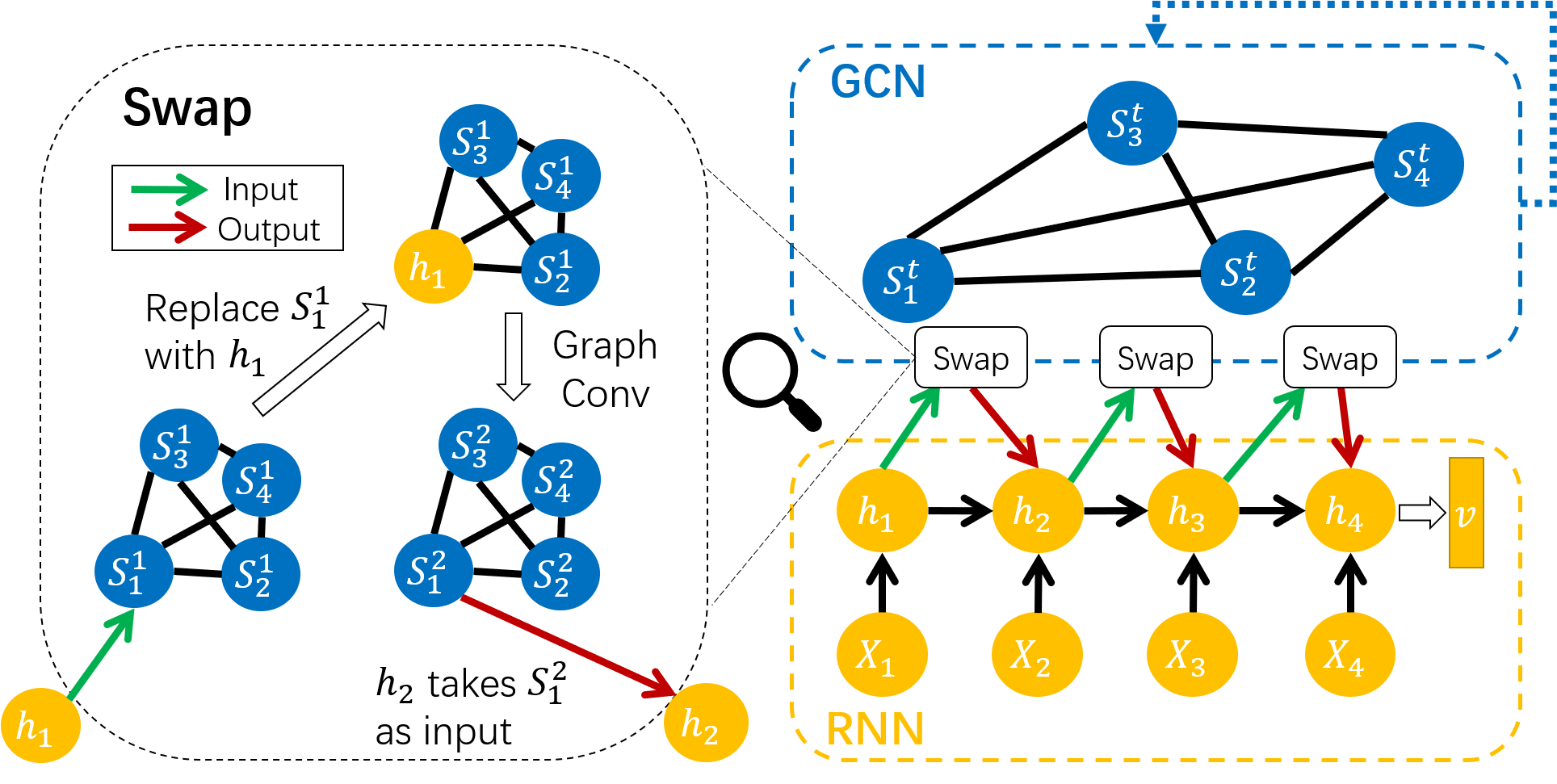}
\caption{RGCN}\label{RGCN&Swap}
\end{subfigure}

\caption{Model architectures. In (a), we demonstrate the pre-processing procedure. We show an example video on how to make hash brown potatoes (YouTube ID: kj5y\_71bsJM). It demonstrates the basic concepts of instructional videos in YouCook2 dataset. \textit{Temporal boundaries} means the human annotated start/end time stamp of a \textit{procedure}, which is well defined in~\cite{Youcook}. Video are segmented into several \textit{segments} (procedures) by the temporal boundaries. Descriptions are also annotated by human, corresponding to each procedure. Transcripts are auto-generated by speech recognition on YouTube. An example QA pair for the video in (a) is, Q:``\textit{How many actions involving physical changes to potatoes are done before adding salt?}'' A:``\textit{2.}''. In (b) and (c), we have question feature attending on each segment. In (d), we illustrate the structure of our proposed RGCN model, where GCN interacts with RNN via ``swap'' operation which takes in RNN's hidden state $h_{t-1}$ and outputs the graph node $S_{t-1}^{t}$ of GCN. We zoom in the first swap operation to provide an intuitive visualization.} 
\label{models}
\end{figure*}

\noindent \textbf{Pre-processing:} \quad 
The videos in our consideration have an average length of 5.27 minutes, which requires us to process the videos into more tractable representations before any sophisticated modelings. Following~\cite{Youcook}, we define \textit{procedure} as the sequence of necessary steps comprising a complex instructional event and segment a video into $N$ procedure segments (see Fig.\ref{preprocessing}). 
To directly benchmark the reasoning ability, we use the ground truth provided by \cite{Youcook} instead to avoid any errors caused by intermediate processing. 
Note that one can apply method developed in \cite{Youcook} for automatically segmentation. 
The frames within each segment are sampled, of which the features are then extracted by ResNet~\cite{ResNet} and encoded by a RNN model. Therefore, we can obtain the features of the procedure segments $\{X_i\}_{i=1}^N\in \mathbb{R}^{d}$ and use them for relation modeling. 

\noindent \textbf{SEQ-SA:} \quad 
We first propose an attention-based RNN model (see Fig.~\ref{SEQ-SA} for an example of $N=4$) to model video representation $v$, where the encoded question feature is used to attend all video features at different time steps. The similarity $a_i$ between question feature $q$ and segment feature $X_i$ is computed by taking the dot product of $q$ and $X_i$: followed by a soft-max normalization: $a_i=\frac{exp(q^{T}X_i)}{\sum exp(q^{T}X_i)}$. Then we multiply each $X_i$ by $a_i$ to obtain the question-attended video feature $X_i^{'}$: $X_i^{'}=a_{i}X_{i}$. Finally, we feed $X_i^{'}$ into an RNN model of which the final hidden state $h_N$ of RNN is taken as the video feature representation $v$.


\noindent \textbf{GCN-SA:} \quad 
We consider a fully-connected graph (see Fig.~\ref{GCN-SA}) to model complex relations among the procedure segments. Although the time dependencies defined by the original video are omitted, different edges in the graph can mine different relations for various reasoning tasks. We use a multi-layer GCN model for this purpose. We define ${\{S_i^{j}\}_{i=1}^N}_{j=1}^M$, where $S_i^{j} \in \mathbb{R}^{d}$, as the graph nodes, where $N$ is the number of nodes within one layer, $M$ is the number of layers. We first initialize nodes $\{S_i^{1}\}_{i=1}^N$ in the first layer by segment features $\{X_i\}_{i=1}^{N}$ correspondingly. We adopt the same GCN structure as described in \cite{GCN}: 
\begin{align}
\mathbf{Z}=\text{ReLU}\{\mathbf{GSW}\}
\enspace,
\label{GCN}
\end{align}
where $\mathbf{G} \in \mathbb{R}^{N\times N}$ represents the adjacency graph, $\mathbf{S} \in \mathbb{R}^{N\times d}$ denotes the concatenation of all node features $\{S_i\}_{i=1}^N$ in one arbitrary layer, and $\mathbf{W} \in \mathbb{R}^{d\times d}$ is the weight matrix which is different for each layer. Each element $G_{ij}$ in $\mathbf{G}$ is the dot product similarity $S_i^T S_j$. Three GCN layers are used in this work, where the output of the previous layer serves as the input of the next layer.

To apply the attention mechanism, we add an additional node in the last layer of the GCN to represent the question feature $q$, and this question node is connected with all other graph nodes $\{S_i^M\}_{i=1}^N$ through $N$ edges. Question node attends to each graph node through different weights on the edges. Similar to SEQ-SA, the weights between $q$ and $\{S_i^M\}_{i=1}^N$ are the dot products of corresponding node pairs, followed by a soft-max normalization. Finally, we use an average pooling operation to compress the output of the last layer $\mathbf{Z} \in \mathbb{R}^{N\times d}$ to $v\in \mathbb{R}^{d}$. 


\noindent \textbf{RGCN:}\quad 
Since the aforementioned GCN-SA is unable to capture the temporal order of video features~\cite{GCN}, and SEQ-SA cannot model the relations between segments with long time spans, we propose a novel Recurrent Graph Convolutional Network (RGCN) architecture (see Fig.~\ref{RGCN&Swap}) to overcome such limitation. The RGCN is a recurrent model that consists of two pathways: RNN and GCN. RNN interacts with GCN mainly through a \textit{swap} operation (see Fig.~\ref{RGCN&Swap}). The details are as follows.

The RNN pathway with $N$ time steps takes in the segment features $X_i$ at each time step. The GCN pathway has $N$ layers, each of which contains $N$ graph nodes. Note that the GCN has the same number of layers as the time steps in RNN pathway. We adopt the same GCN architecture as described in GCN-SA model except that a recurrent computation paradigm is applied here, where the weights $\mathbf{W}$ is shared among all layers. The computation within the RNN memory cell at each time step and the computation of each GCN layer are performed alternatively.
For each time step $t$, we first concatenate together the segment feature $X_t$ and the feature of node $S^{t}_{t-1}$ in GCN, which is then used as the input to RNN memory cell at the $t$-th time step. Following~\cite{LSTM}, we update the hidden state $h_t$ of RNN:
\begin{align}
h_t=\text{RNN}\{[X_t,S^{t}_{t-1}],h_{t-1}\}
\enspace,
\label{RNN}
\end{align}

Then we replace GCN's graph node $S^{t}_{t}$ with the updated hidden state $h_t$ of RNN. This \textit{swap} operation act as a bridge between RNN and GCN for message passing. Finally, the $(t+1)$-th GCN layer takes all $\{S^t_i\}_{i=1}^N$ as input to compute the response $\{S^{t+1}_i\}_{i=1}^N$:
\begin{align}
\mathbf{Z_{t+1}}=\text{ReLU}\{\mathbf{GZ_{t}W}\}
\enspace,
\label{RGCN_GCN}
\end{align}
where $\mathbf{Z_{t}}$ is the concatenation of $\{S_i^t\}_{i=1}^N$. We take the final hidden state $h_N$ of RNN as the video representation $v$.

Additionally, we extend the proposed RGCN with attention mechanism. The two pathways corresponds to the SEQ and GCN model, so we simply adopt how attention is cast on both pathways, and obtain RGCN-SA.


\subsection{Multiple modalities} 
\label{sec:modalities}
Besides videos and questions, we further investigate how much benefit we can obtain from other modalities such as narratives, which is very common in instructional videos. We are interested in two types of narratives, namely transcripts and descriptions.

\textbf{Transcripts:} The audio signal is an important modality for videos. In our dataset, the valuable audio information in videos is all chefs speaking. Therefore, we substitute audio with auto-generated transcripts on YouTube. Transcripts, which can be seen as describing the corresponding procedures, are highly unstructured, noisy, and misaligned narratives~\cite{visual-linguistic} in that chefs may talk about things not related to the cooking procedure, or that the speech recognition on YouTube may generate some unexpected sentences. Nevertheless, it can provide extra information to solve visual ambiguities, e.g., distinguishing water from white vinegar, which both look transparent.

\textbf{Descriptions:} In YouCook2 dataset, each procedure in a video corresponds to a sentence of natural language description annotated by a human. Different from transcripts, descriptions are much less dense with respect to time, and can be seen as highly constrained narratives because human labor is applied to extract the essence of the corresponding procedures. Each piece of description is associated with the procedure it describes because they are highly related semantically .

For each individual modality (which can be description or transcripts), we aim to model a feature representation $m$, then fuse it with $v$ and $q$ to predict the answer $A^*$. To achieve this goal, we make use of a hierarchical RNN structure: a lower-level RNN models the natural language words within each segment, and a higher level RNN models the gloabal feature of the video.

\section{Experiments}
\label{sec:experiment}
First, we introduce the implementation details of the training process. Then some baseline models are described, followed by results analysis. Also, we explored the benefit introduced by other modalities such as description and transcripts. All experiments conducted in this work are evaluated on both multiple choice and K-Space evaluation metrics. In Tab.~\ref{segment-results}, only multiple choice accuracy is provided for discussion. All other results on K-Space are in supplementary material.

\begin{table*}[]
\centering
\caption{Results on different model architectures.}
\begin{tabular}{l|llllll|l}
\hline
                      & Count          & Order          & Taste          & Time           & Complex        & Property       & All            \\ \hline
\textbf{Common sense} & 0.535          & 0.432           & 0.654          & 0.485          & 0.511          & 0.588          & 0.528          \\ \hline\hline
Bare QA               & 0.435          & 0.321          & 0.466          & 0.239          & 0.292          & 0.438          & 0.348          \\ \hline
Naive RNN            & 0.434           & 0.330          & 0.467          & 0.234          & 0.283          & 0.449          & 0.347          \\ \hline
MAC                   & 0.438          & 0.331          & 0.462          & 0.229          & 0.294          & 0.437          & 0.348          \\ \hline\hline
SEQ                   & 0.452          & 0.337          & 0.476          & 0.230          & 0.288          & 0.449          & 0.352          \\ \hline
GCN                   & 0.452          & 0.341          & 0.464          & 0.224          & 0.282          & 0.427          & 0.346          \\ \hline
\textbf{RGCN}         & \textbf{0.522} & \textbf{0.371} & 0.478          & \textbf{0.277} & \textbf{0.329} & \textbf{0.490} & \textbf{0.392} \\ \hline\hline
SEQ-SA               & 0.473          & 0.355          & \textbf{0.483} & 0.256          & 0.316          & 0.465          & 0.373          \\ \hline
GCN-SA               & 0.477          & 0.343          & \textbf{0.487} & 0.229          & 0.311          & 0.446          & 0.365          \\ \hline
\textbf{RGCN-SA}     & \textbf{0.545} & \textbf{0.367} & 0.481          & \textbf{0.279} & \textbf{0.316} & \textbf{0.486} & \textbf{0.403} \\ \hline
\end{tabular}
\label{segment-results}
\end{table*}

\subsection{Implementation details}
\label{sec:implementation}
Our codes are based on PyTorch deep learning framework. ResNet is used to extract visual features of 500 frames in each video, producing a 512-d vector. By using embedding layers, the question words are transformed into 300-d vectors which are optimized during the training process. For all models involving RNNs in this work, we apply single direction LSTMs~\cite{LSTM} (an improved version of vanilla RNN) with 512 hidden units. Adam optimizer is used with the learning rate of 0.0001. 

We split the training/testing set according to the original YouCook2 dataset. All videos in the YouCook2 training set are used as training videos in our dataset. Therefore, there are 10,179 QA pairs in our training set, and the rest are treated as testing set.

\subsection{Baselines}
\label{sec:exp:alter}

We set up some baseline models which takes no instructional information. In other words, only the original video is presented to the models without temporal boundaries or descriptions.

\noindent \textbf{Bare QA:} \quad 
First, we build the QA model which predicts answers based on questions only (without videos). Then for multiple choice, the answer is predicted by a similar way as Eq.~\ref{f_vqa_mc}: $j^{*}=\argmax_{j=1\dots M} f(q,a)$. For K-Space, we adopt a similar formula as Eq.~\ref{f_vqa_ks}: $A^{*}=\argmax_{j=1,\dots,K} g_j(q)$.

\noindent \textbf{Naive RNN:} \quad 
RNN is a base of most state-of-the-art ImageQA~\cite{image_QA_1,image_QA_2,image_QA_3} and VideoQA\cite{gradually,leveraging} models. Instead of applying the segmentation pre-processing which we introduced in Sec.~\ref{sec:models}, Naive RNN takes in the ResNet feature of sampled video frames directly. Similar to other models discussed previously, we take the final hidden state of the RNN as the video feature $v$. Then we evaluate the model performance based on Eq.~\ref{f_vqa_mc} and Eq.~\ref{f_vqa_ks}. 

\noindent \textbf{MAC:} \quad 
MAC~\cite{MAC} is currently the state-of-the-art model on CLEVR dataset. Since our proposed YouQuek dataset shares similar question style with CLEVR dataset, we adopt MAC as another alternative model. To apply MAC which is designed for spatial reasoning to the temporal reasoning task in our work, we replace the input image features $\{I_i\}_{i=1}^L$, where $I_i\in \mathbb{R}^{d}$ ($L$ is the number of spatial dimension of an image), with video frame features $\{X_i\}_{i=1}^N$, where $X_i\in \mathbb{R}^{d}$ ($N$ is the number of sampled frames).

\noindent \textbf{Human quiz:} \quad
Apart from using deep learning models to complete VideoQA tasks, we also invite ten human annotators to perform human test. First, they are asked to answer the questions without any other information, but by guessing or using common sense. Second, they are allowed to watch the videos without audio. Finally, audio is also turned on to correspond with transcripts. Details of the setting are in supplementary material.

\subsection{Results Analysis}
\label{sec:results}
\begin{table*}[!ht]
\centering
\caption{Results on multiple modalities, where V stands for visual information, CC for transcripts, and D for descriptions.} 
\begin{tabular}{l||ll||ll||ll||ll||ll||ll}
\hline
\multirow{2}{*}{} & \multicolumn{2}{l||}{\textbf{SEQ}} & \multicolumn{2}{l||}{\textbf{SEQ-SA}} & \multicolumn{2}{l||}{\textbf{GCN}} & \multicolumn{2}{l||}{\textbf{GCN-SA}} & \multicolumn{2}{l||}{\textbf{RGCN}}  & \multicolumn{2}{l}{\textbf{RGCN-SA}} \\ \cline{2-13} 
                  & MC             & KS             & MC             & KS            & MC           & KS          & MC              & KS                                      & MC          & KS                   & MC          & KS                  \\ \hline
Visual            & 0.352          & \textbf{0.160} & 0.373          & 0.164         & 0.346        & 0.150       & 0.365           & 0.164                                   & 0.392       & 0.179                & 0.403       & 0.182               \\ \hline
CC                & 0.346          & 0.159          & 0.353          & 0.152         & 0.343        & 0.143       & 0.346           & 0.150                                   & 0.361       & 0.152                & 0.366       & 0.144               \\ \hline
Description       & \textbf{0.353} & 0.158          & 0.365          & 0.156         & \textbf{0.352}& \textbf{0.157}& 0.347        & 0.153                                   & 0.385       & 0.163                & 0.389       & 0.162               \\ \hline
V+CC              & 0.347          & 0.151          & 0.375          & 0.167         & 0.348        & 0.150       & 0.375           & 0.177                                   & 0.390       & 0.173                & 0.393       & 0.180               \\ \hline
V+D               & 0.351          & \textbf{0.160} & \textbf{0.379} & \textbf{0.173}& 0.349        & 0.148       & \textbf{0.383} & \textbf{0.183}                           & \textbf{0.413} & \textbf{0.194}    & \textbf{0.416} & \textbf{0.203}   \\ \hline
\end{tabular}
\label{multimodalities}
\end{table*}

Tab.~\ref{segment-results} shows the experiment results on all models and baselines. 
We start with the comparison among baseline models that are without temporal boundary information (i.e., Bare QA, Naive RNN and MAC). As we can see from row 2 to row 4 of Tab.~\ref{segment-results} that the three baselines have very close overall accuracy. Though Naive RNN take in the video stream, it cannot achieve better results than the bare QA. Therefore, we claim that as the base of most state-of-the-art visual QA models, RNN fails to extract meaningful visual information for instructional video reasoning. The reason is that it is difficult for RNN to model complex relations due to its sequential structure. Another reason is that RNN cannot capture long time dependencies of videos due to the memory limitation, even for RNN variants such as LSTM and GRU. As the best model on CLEVR, MAC achieves the same overall accuracy with Bare QA on YouQuek, which demonstrates the special difficulty of video understanding compared with ImageQA  

Then we analyze the performance of models proposed in Sec~\ref{sec:models}, which incorporate temporal boundary information of instructional videos to boost the performance. Recall that the temporal boundaries are provided by the ground truth in~\cite{Youcook}. First, to evaluate the improvement introduced by attention mechanism, we remove the question attention operation to formulate the models: SEQ, GCN, RGCN, the results of which are shown in row 5 to row 7 of Tab.~\ref{segment-results}. We can see from row 5 to row 10 of Tab.~\ref{segment-results} that the margins gained by introducing attention are from 1.1\% to 2.1\%, which demonstrates that question can guide the models to extract more meaningful features, and all these models outperform baselines by a big margin. Especially, RGCN-SA achieves the highest overall accuracy of 40.3\%, 5.5\% higher than MAC, and SEQ-SA ranks second among the attention based models with an overall accuracy of 37.3\%. This demonstrates that the procedure segmentation helps models make better use of video streams. 

Finally, we investigate the performance of attention based models on various question categories. The comparison between SEQ-SA and GCN-SA shows that GCN-SA achieves higher accuracy scores on ``count'' and ``taste'' questions, while SEQ-SA performs better on all other categories. Intuitively, ``order'', ``property'' questions require temporal order information to be answered, because the questions usually contain sequence-related keywords, e.g., ``before/after/between''. Graph structure can hardly capture such ordering information. Nevertheless, the capability of modeling relations gives graph structure a reasonably good performance, especially on ``count'' and ``taste'' questions which challenge less on ordering. Since both sequence and graph models show advantages on different categories of questions, we take the advantages of both two models to build RGCN-SA, which is capable of passing messages between the two different pathways. Results show that graph and sequence can boost each others' performance on most question types except for ``taste''.

\subsection{Multimodalities}
Based on temporal boundary annotations, we further explore other modalities. As described in Sec.~\ref{sec:modalities}, we experiment on two types of narratives, unconstrained transcripts and concentrated descriptions. Descriptions are already associated with video segments in the YouCook2 dataset, so we only need to align the transcripts with segments by selecting transcripts that lay between the temporal boundaries. Results are shown in Tab.~\ref{multimodalities}.

As for different modalities, we first compare visual information, transcripts and description separately. Although descriptions are human annotated, highly refined reconstruction of the content of instructional videos, mere description seems not helpful when compared with visual information. Transcripts, to be worse, always decrease the performance. However, when narratives and visual information are combined together, we can see a significant increase in accuracy scores. SEQ-SA, GCN-SA, RGCN and RGCN-SA all achieve highest multiple choice accuracy when trained with both visual features and descriptions. SEQ with visual and description information also gets the highest K-Space accuracy compared to SEQ models trained on other modalities. However, transcripts still fail to provide as much valuable information as descriptions on videos, since the performance of models with visual and transcript information is worse than visual plus description. Transcripts even have a negative effect on SEQ and RGCN in that multiple choice accuracy is dropped when transcripts are added to visual information. Possible reasons are that the transcripts are too dense, and the quality of auto-generated transcripts are uncontrollable. As for different structures, we can see that our RGCN-SA still achieves the highest performance, while all attention models provides reasonable results.

\subsection{Human quiz}
\label{sec:human-quiz}
In the human quiz part, participants are asked to do three sets of tests, namely guessing with common sense, with visual information, and with both visual and audio information. The results of the guessing step are shown at the top row in Tab.~\ref{segment-results}. As we can see, even without any video information, human can achieve an accuracy as high as $52.8\%$. An interesting fact here is that human participants did a good job on the ``when'' questions, which is unexpected because one cannot know the exact time point of what is going to happen without watching the video. The reason is that humans have an intuition of which ingredients is more likely to be added first, or which step is less likely to happen at the beginning, owing to their common sense or life experience. Another support for the power of common sense is the high accuracy score for ``taste'' questions. For machines, the taste can only possibly be learned from the relations between ingredients and correct answers. However, for human beings, the tastes of different ingredients is already known in daily life. Given visual information, the human performance becomes almost perfect ($97.0\%$), so the accuracy scores are not provided in the form of tables. This is reasonable because human has a powerful visual understanding and comprehending system. Given that the accuracy is already very high and that the dataset is collected without audio information, the improvement is minor ($97.7\%$) after adding audio information. It is worth mention that RGCN-SA outperforms the human baseline on ``count'' questions, yet there is still a long way to go in visual reasoning tasks.

\section{Conclusion}
\label{sec:conclusion}
In this paper, we emphasize reasoning on instructional videos. We construct YouCook Question Answering (YouQuek) dataset, and three models with sequence (SEQ), graph (GCN), and fused (recurrent graph convolutional network, RGCN) structures are proposed to explore the instructional information. Attention mechanism is applied on the proposed models to boost performance, and RGCN-SA achieves the best accuracy on both multiple choice and K-Space evaluation metrics. Experiment results show that the proposed RGCN successfully fuse the order and relation information together for modeling instructional videos. Also, multiple modalities for instructional videos are analyzed, showing that human annotated temporal boundaries and descriptions are critical for instructional video reasoning.

\section*{Acknowledgement}
This work was supported in part by NSF IIS-1813709, IIS-1741472, and CHE-1764415. Any opinions, findings, and conclusions or recommendations expressed in this material are those of the authors and do not necessarily reflect the views of the NSF.

{\small
\bibliographystyle{ieee}
\bibliography{egbib}

\begin{thebibliography}{10}\itemsep=-1pt

\bibitem{changing-tires-cvpr}
J.~Alayrac, P.~Bojanowski, N.~Agrawal, J.~Sivic, I.~Laptev, and
  S.~Lacoste{-}Julien.
\newblock Unsupervised learning from narrated instruction videos.
\newblock In {\em 2016 {IEEE} Conference on Computer Vision and Pattern
  Recognition, {CVPR} 2016, Las Vegas, NV, USA, June 27-30, 2016}, pages
  4575--4583, 2016.

\bibitem{image_QA_1}
S.~Antol, A.~Agrawal, J.~Lu, M.~Mitchell, D.~Batra, C.~L. Zitnick, and
  D.~Parikh.
\newblock {VQA:} visual question answering.
\newblock In {\em {ICCV}}, pages 2425--2433. {IEEE} Computer Society, 2015.

\bibitem{ResNet}
K.~He, X.~Zhang, S.~Ren, and J.~Sun.
\newblock Deep residual learning for image recognition.
\newblock In {\em 2016 {IEEE} Conference on Computer Vision and Pattern
  Recognition, {CVPR} 2016, Las Vegas, NV, USA, June 27-30, 2016}, pages
  770--778, 2016.

\bibitem{good-question}
M.~Heilman and N.~A. Smith.
\newblock Good question! statistical ranking for question generation.
\newblock In {\em Human Language Technologies: Conference of the North American
  Chapter of the Association of Computational Linguistics, Proceedings, June
  2-4, 2010, Los Angeles, California, {USA}}, pages 609--617, 2010.

\bibitem{LSTM}
S.~Hochreiter and J.~Schmidhuber.
\newblock Long short-term memory.
\newblock {\em Neural Computation}, 9(8):1735--1780, 1997.

\bibitem{visual-linguistic}
D.~Huang, J.~J. Lim, L.~Fei{-}Fei, and J.~C. Niebles.
\newblock Unsupervised visual-linguistic reference resolution in instructional
  videos.
\newblock In {\em 2017 {IEEE} Conference on Computer Vision and Pattern
  Recognition, {CVPR} 2017, Honolulu, HI, USA, July 21-26, 2017}, pages
  1032--1041, 2017.

\bibitem{finding-it}
D.-A. Huang, S.~Buch, L.~M. Dery, A.~Garg, L.~Fei-Fei, and J.~C. Niebles.
\newblock Finding “ it ” : Weakly-supervised reference-aware visual
  grounding in instructional videos.
\newblock 2018.

\bibitem{MAC}
D.~A. Hudson and C.~D. Manning.
\newblock Compositional attention networks for machine reasoning.
\newblock {\em CoRR}, abs/1803.03067, 2018.

\bibitem{CLEVR}
J.~Johnson, B.~Hariharan, L.~van~der Maaten, L.~Fei{-}Fei, C.~L. Zitnick, and
  R.~B. Girshick.
\newblock {CLEVR:} {A} diagnostic dataset for compositional language and
  elementary visual reasoning.
\newblock In {\em 2017 {IEEE} Conference on Computer Vision and Pattern
  Recognition, {CVPR} 2017, Honolulu, HI, USA, July 21-26, 2017}, pages
  1988--1997, 2017.

\bibitem{image_QA_3}
K.~Kafle and C.~Kanan.
\newblock Answer-type prediction for visual question answering.
\newblock In {\em {CVPR}}, pages 4976--4984. {IEEE} Computer Society, 2016.

\bibitem{making-coffee-cvpr}
H.~Kuehne, A.~B. Arslan, and T.~Serre.
\newblock The language of actions: Recovering the syntax and semantics of
  goal-directed human activities.
\newblock In {\em 2014 {IEEE} Conference on Computer Vision and Pattern
  Recognition, {CVPR} 2014, Columbus, OH, USA, June 23-28, 2014}, pages
  780--787, 2014.

\bibitem{image_QA_2}
M.~Malinowski and M.~Fritz.
\newblock A multi-world approach to question answering about real-world scenes
  based on uncertain input.
\newblock In {\em Advances in Neural Information Processing Systems 27: Annual
  Conference on Neural Information Processing Systems 2014, December 8-13 2014,
  Montreal, Quebec, Canada}, pages 1682--1690, 2014.

\bibitem{MPI}
M.~Rohrbach, S.~Amin, M.~Andriluka, and B.~Schiele.
\newblock A database for fine grained activity detection of cooking activities.
\newblock In {\em {CVPR}}, pages 1194--1201. {IEEE} Computer Society, 2012.

\bibitem{video-clevr}
X.~Song, Y.~Shi, X.~Chen, and Y.~Han.
\newblock Explore multi-step reasoning in video question answering.
\newblock In {\em 2018 {ACM} Multimedia Conference on Multimedia Conference,
  {MM} 2018, Seoul, Republic of Korea, October 22-26, 2018}, pages 239--247,
  2018.

\bibitem{MovieQA}
M.~Tapaswi, Y.~Zhu, R.~Stiefelhagen, A.~Torralba, R.~Urtasun, and S.~Fidler.
\newblock Movieqa: Understanding stories in movies through question-answering.
\newblock In {\em 2016 {IEEE} Conference on Computer Vision and Pattern
  Recognition, {CVPR} 2016, Las Vegas, NV, USA, June 27-30, 2016}, pages
  4631--4640, 2016.

\bibitem{video-captioning}
S.~Venugopalan, M.~Rohrbach, J.~Donahue, R.~J. Mooney, T.~Darrell, and
  K.~Saenko.
\newblock Sequence to sequence - video to text.
\newblock In {\em {ICCV}}, pages 4534--4542. {IEEE} Computer Society, 2015.

\bibitem{GCN}
X.~Wang and A.~Gupta.
\newblock Videos as space-time region graphs.
\newblock In {\em {ECCV} {(5)}}, volume 11209 of {\em Lecture Notes in Computer
  Science}, pages 413--431. Springer, 2018.

\bibitem{visual-grounding}
F.~Xiao, L.~Sigal, and Y.~J. Lee.
\newblock Weakly-supervised visual grounding of phrases with linguistic
  structures.
\newblock In {\em {CVPR}}, pages 5253--5262. {IEEE} Computer Society, 2017.

\bibitem{gradually}
D.~Xu, Z.~Zhao, J.~Xiao, F.~Wu, H.~Zhang, X.~He, and Y.~Zhuang.
\newblock Video question answering via gradually refined attention over
  appearance and motion.
\newblock In {\em Proceedings of the 2017 {ACM} on Multimedia Conference, {MM}
  2017, Mountain View, CA, USA, October 23-27, 2017}, pages 1645--1653, 2017.

\bibitem{Unifying}
H.~Xue, Z.~Zhao, and D.~Cai.
\newblock Unifying the video and question attentions for open-ended video
  question answering.
\newblock {\em {IEEE} Trans. Image Processing}, 26(12):5656--5666, 2017.

\bibitem{leveraging}
K.~Zeng, T.~Chen, C.~Chuang, Y.~Liao, J.~C. Niebles, and M.~Sun.
\newblock Leveraging video descriptions to learn video question answering.
\newblock In {\em Proceedings of the Thirty-First {AAAI} Conference on
  Artificial Intelligence, February 4-9, 2017, San Francisco, California,
  {USA.}}, pages 4334--4340, 2017.

\bibitem{spatio-temopral}
Z.~Zhao, Q.~Yang, D.~Cai, X.~He, and Y.~Zhuang.
\newblock Video question answering via hierarchical spatio-temporal attention
  networks.
\newblock In {\em Proceedings of the Twenty-Sixth International Joint
  Conference on Artificial Intelligence, {IJCAI} 2017, Melbourne, Australia,
  August 19-25, 2017}, pages 3518--3524, 2017.

\bibitem{Youcook}
L.~Zhou, C.~Xu, and J.~J. Corso.
\newblock Towards automatic learning of procedures from web instructional
  videos.
\newblock In {\em Proceedings of the Thirty-Second {AAAI} Conference on
  Artificial Intelligence, (AAAI-18), the 30th innovative Applications of
  Artificial Intelligence (IAAI-18), and the 8th {AAAI} Symposium on
  Educational Advances in Artificial Intelligence (EAAI-18), New Orleans,
  Louisiana, USA, February 2-7, 2018}, pages 7590--7598, 2018.

\bibitem{dense-captioning}
L.~Zhou, Y.~Zhou, J.~J. Corso, R.~Socher, and C.~Xiong.
\newblock End-to-end dense video captioning with masked transformer.
\newblock {\em CoRR}, abs/1804.00819, 2018.

\bibitem{uncovering}
L.~Zhu, Z.~Xu, Y.~Yang, and A.~G. Hauptmann.
\newblock Uncovering temporal context for video question and answering.
\newblock {\em CoRR}, abs/1511.04670, 2015.

\end{thebibliography}
}

\end{document}